\documentclass[letterpaper, 10 pt, conference]{ieeeconf}  

\IEEEoverridecommandlockouts                              
                                                        
\usepackage{amsmath} 
\usepackage{amssymb}  
\usepackage[usenames, dvipsnames]{color}
\usepackage[short]{optidef}
\usepackage{mathtools}
\usepackage{algorithm}
\usepackage{algorithmicx}
\usepackage{pgfplots}
\usepackage{graphicx}
\usepackage[abs]{overpic}
\usepackage{booktabs}
\usepackage{caption}
\usepackage{algpseudocode}
\usepackage{hyperref}
\usepackage{bm}
\usepackage{siunitx}
\usepackage{threeparttable}
\usepackage{makecell}
\usepackage[table]{xcolor} 
\usepackage{adjustbox}
\usepackage{pifont}
\usepackage{pgfplots}
\usepackage{empheq}
\usepackage{cuted}
\usepackage{stfloats}
\usepackage{nicematrix, xcolor}

\usetikzlibrary{arrows.meta}
\usetikzlibrary{backgrounds}
\usetikzlibrary{plotmarks}
\usepgfplotslibrary{patchplots}
\usepgfplotslibrary{fillbetween}
\pgfplotsset{%
    layers/standard/.define layer set={%
        background,axis background,axis grid,axis ticks,axis lines,axis tick labels,pre main,main,axis descriptions,axis foreground%
    }{
        grid style={/pgfplots/on layer=axis grid},%
        tick style={/pgfplots/on layer=axis ticks},%
        axis line style={/pgfplots/on layer=axis lines},%
        label style={/pgfplots/on layer=axis descriptions},%
        legend style={/pgfplots/on layer=axis descriptions},%
        title style={/pgfplots/on layer=axis descriptions},%
        colorbar style={/pgfplots/on layer=axis descriptions},%
        ticklabel style={/pgfplots/on layer=axis tick labels},%
        axis background@ style={/pgfplots/on layer=axis background},%
        3d box foreground style={/pgfplots/on layer=axis foreground},%
    },
}

\newcommand{\na}{\textemdash}
\newcommand{\cmark}{\ding{51}}%
\newcommand{\xmark}{\ding{55}}%

\DeclareMathOperator*{\argmin}{arg\,min}

\usepackage{cite}

\newif\ifanonymous
\anonymousfalse

\title{\LARGE \bf
Complementarity by Construction: A Lie-Group Approach to Solving Quadratic Programs with Linear Complementarity Constraints
}

\ifanonymous
\author{
Authors omitted for review
}
\else
\author{
Arun L. Bishop$^{1}$, Micah I. Reich$^{1}$, and Zachary Manchester$^{2}$
\thanks{$^{1}$Authors are with the Robotics Institute, Carnegie Mellon University}%
\thanks{$^{2}$Author is with the Department of Aeronautics and Astronautics, Massachusetts Institute of Technology}%
}
\fi

\pgfplotsset{}
\begin{document}

\makeatletter
\let\@oldmaketitle\@maketitle
\renewcommand{\@maketitle}{%
  \@oldmaketitle
  \centering
  \includegraphics[width=0.9\linewidth]{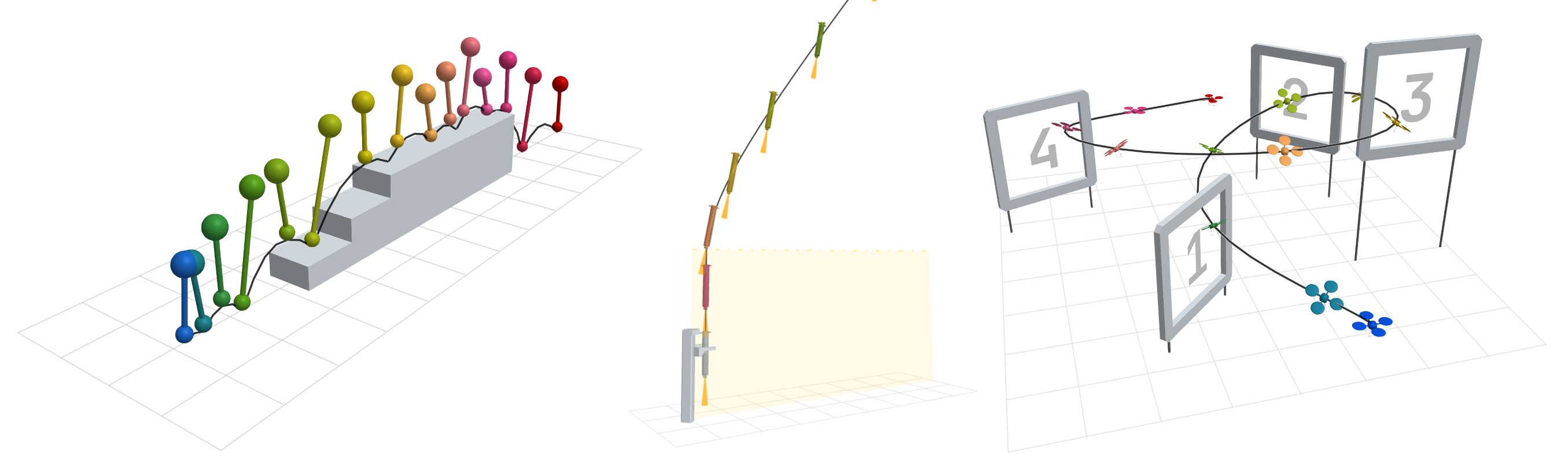}
  \captionof{figure}{
    Quadratic programs with linear complementarity constraints can model a wide variety of robotics problems with non-smooth or ``switching'' constraints. Here we show a hopper hopping over a raised platform with contact and friction constraints (left), a rocket catch with state-triggered orientation and keep-out constraints to avoid collision and plume impingement on the catch tower (middle), and a quadrotor flying through racing gates with completion order constraints (right).
  }
  \label{fig:top_figure}
  \vspace{-5mm}
}
\makeatother

\let\oldmaketitle\maketitle
\renewcommand{\maketitle}{%
  \oldmaketitle
  \addtocounter{figure}{-1}  
}

\maketitle
\thispagestyle{empty}
\pagestyle{empty}

\begin{abstract}
Many problems in robotics require reasoning over a mix of continuous dynamics and discrete events, such as making and breaking contact in manipulation and locomotion. These problems are locally well modeled by linear complementarity quadratic programs (LCQPs), an extension to QPs that introduce complementarity constraints. While very expressive, LCQPs are non-convex, and few solvers exist for computing good local solutions for use in planning pipelines. In this work, we observe that complementarity constraints form a Lie group under infinitesimal relaxation, and leverage this structure to perform on-manifold optimization. We introduce a retraction map that is numerically well behaved, and use it to parameterize the constraints so that they are satisfied by construction. The resulting solver avoids many of the classical issues with complementarity constraints. We provide an open-source solver, Marble, that is implemented in C++ with Julia and Python bindings. We demonstrate that Marble is competitive on a suite of benchmark problems, and solves a number of robotics problems where existing approaches fail to converge.

\end{abstract}

\section{INTRODUCTION}
Many important planning and control problems in robotics involve optimizing over a mix of continuous and discrete elements. For example, manipulation and locomotion extend smooth motion planning to include decisions over when to make and break contact with the environment. Problems with binary goals or region-specific constraints like those in Figure 1 also introduce discrete decision variables \cite{szmuk_successive_2020, dempe_mpec_2020}. Solving these problems globally is known to be NP-hard and combinatorially expensive, but in many cases we are satisfied with local feasible solutions, such as in model-predictive control or as a heuristic in planning pipelines.

It is well established that smooth nonlinear optimization problems are locally well modeled by quadratic programs (QPs) \cite{nocedal_numerical_2006}. In the context of nonlinear trajectory optimization, contact-implicit methods model the physics of contact through complementarity constraints \cite{stewart_implicit_1996}. In these cases, a natural choice for local approximations are linear complementary quadratic programs (LCQPs). LCQPs extend standard QPs with linear complementarity constraints: $s \geq 0$, $t \geq 0$, and $s \odot t = 0$, also denoted $0 \leq s \perp t \geq 0$. LCQPs are very expressive, and are capable of capturing the local behavior of many non-smooth systems: Bi-level optimization, binary problems, and game-theoretic problems can all be modeled with LCQPs \cite{ferris_engineering_1997}. LCQPs are, however, non-convex and challenging to solve, and there has been relatively little research focused on developing fast local solvers that leverage the specific structure of these problem \cite{chen_class_2009, bai_convex_2013, ralph_c-index_2011}. 

In this work, we present a novel approach for solving LCQPs that addresses many of the challenges of optimizing over complementarity constraints. Our approach solves a series of subproblems with relaxed complementarity constraints similar to \cite{howell_calipso_2023}, but is the first to take advantage of the smooth manifold structure of the feasible set. Our work is inspired by \cite{permenter_log-domain_2023}, which introduced a log-domain parameterization of relaxed complementarity constraints through an exponential retraction map for use in interior-point methods for convex QPs. Our contributions are:
\begin{itemize}
    \item A parameterization of relaxed complementarity that uses its Lie group structure to satisfy the constraints by construction
    \item A proposed retraction map that avoids the numerical ill-conditioning of the exponential map
    \item A corresponding LCQP solver with an open-source C++ implementation with Julia and Python bindings
    \item Comparisons to LCQPow, a state-of-the-art LCQP solver on a standard benchmark suite and a variety of robotics problems, some of which LCQPow fails to solve, showing improved performance
\end{itemize}

This paper is organized as follows. Section \ref{sec:background} presents the necessary mathematical background, and Section \ref{sec:existing_approaches} reviews existing approaches for solving LCQPs. Section \ref{sec:comp} introduces our reformulation and retraction map, while Section \ref{sec:coco} describes the proposed solver. Section \ref{sec:experiments} presents the experimental results, and Section \ref{sec:conclusions} concludes with a discussion of limitations and directions for future work.


\section{Background}\label{sec:background}

\subsection{Complementarity Constraints}\label{sec:comp}
Complementarity constraints take the following form over vectors $s \in \mathbb{R}^p$ and $t \in \mathbb{R}^p$ where $\odot$ denotes element-wise multiplication:
\begin{align}
    \label{eq:complementarity_definition}
    0 \leq s \perp t \geq 0 \iff 
    \begin{cases}
        s \geq 0,  \,\, t \geq 0 \\ s \odot t = 0
    \end{cases}
\end{align}
The L-shaped feasible region for \eqref{eq:complementarity_definition} is both non-convex and non-smooth at the corner $(s, t) = (0, 0)$, as shown in Figure \ref{fig:retraction}. When combined with other linear constraints, disjoint feasible regions can occur. In addition, complementarity constraints violate the linear independence constraint qualification (LICQ): At any solution $\bar{s} = 0, \bar{t} > 0$, the gradients of  the active constraints are linearly dependent:
\begin{align*}
    \nabla(s)\Big|_{(\bar{s}, \bar{t})} = \begin{pmatrix}
        1 \\ 0
    \end{pmatrix}, \quad
    \nabla (st)\Big|_{(\bar{s}, \bar{t})} = \begin{pmatrix}
        \bar{t} \\ 0
    \end{pmatrix}
\end{align*}
LICQ is often a required condition for convergence of general nonlinear program (NLP) solvers as it ensures uniqueness of the Lagrange multipliers \cite{luo_mathematical_1996}.

\subsection{Linear-Complementarity Quadratic Programs}\label{sec:lcqp} 

LCQPs are optimization problems that minimize a quadratic cost subject to linear equality, inequality, and complementarity constraints:
\begin{subequations}
\label{eq:original_problem_formulation}
\begin{align}
    \min_{z, s, t} \quad &\frac{1}{2} z^\top Qz+g^\top z \label{eq:LCQP_cost} \\
    \text{subject to} \quad &Az + b \geq 0
    \label{eq:original_linear_ineq} \\
    &Lz + l = s
\label{eq:original_complementarity_left} \\
    &Rz + r = t
\label{eq:original_complementarity_right} \\
    &0 \leq s \perp t \geq 0 \label{eq:original_complementarity}
\end{align}
\end{subequations}
Here, $z \in \mathbb{R}^n$ is the solution vector, $A \in \mathbb{R}^{m\times n}$ and $L, R \in \mathbb{R}^{p\times n}$ are the constraint Jacobians for the inequality and complementarity constraints, respectively, $b \in \mathbb{R}^{m}$ and $l, r \in \mathbb{R}^{p}$ are the affine terms for each constraint, and $s, t \in \mathbb{R}^{p}$ are complementarity slack variables. A standard QP is comprised of the cost \eqref{eq:LCQP_cost} and constraints \eqref{eq:original_linear_ineq}, but the difficulty of solving \eqref{eq:original_problem_formulation} comes from the additional constraints \eqref{eq:original_complementarity_left}--\eqref{eq:original_complementarity} as discussed in Section \ref{sec:comp}.

\subsection{Lie Groups and Lie Algebras}

Lie groups are groups that also have a smooth manifold structure (i.e. they are continuous). Common examples in robotics include the 2D and 3D rotation groups, $SO(2)$ and $SO(3)$, the group of 3D rigid-body motions $SE(3)$, and the unit quaternions $SU(2)$. An accessible introduction for roboticists can be found in \cite{sola_micro_2021}. Lie groups must be closed under a multiplication operation, and must have an identity element and inverse. Importantly for us, there is a very well-developed theory of optimization on Lie groups that takes advantage of their differentiability and algebraic structure \cite{absil_optimization_2009}.

While Lie groups are not vector spaces, they can be locally linearized to enable vector-space calculations in optimization algorithms. Roughly speaking, calculations involving gradients, Jacobians, and Hessians can be performed on the Lie algebra, which is the linearization (i.e. tangent space) of a Lie group at the identity \cite{stillwell_naive_2008}. Vectors in this tangent space can then be mapped back onto the group via a retraction map \cite{sola_micro_2021}, enabling many familiar algorithms like Newton's method \cite{absil_optimization_2009}\cite{jackson_planning_2021} and Kalman filters \cite{barrau_invariant_2018} to be easily ported. A major benefit of this approach is that algorithms operating in the Lie algebra do not need to explicitly reason about manifold constraints: the retraction map ensures that all iterates stay on the group manifold by construction.

Perhaps the simplest illustrative example of a Lie group is the positive real numbers $\mathbb{R^+}$, with standard scalar multiplication and the identity element 1. The corresponding Lie algebra is the real numbers $\mathbb{R}$, and the standard retraction map is the exponential, i.e. every positive number $x$ can be written as $e^a$ for some scalar $a$. By calculating optimization steps on the Lie algebra $\mathbb{R}$ and applying the exponential map, iterates are guaranteed to stay positive, and we do not need to explicitly enforce the constraint $x \geq 0$.

\subsection{Interior-point and Barrier Methods} \label{sec:interior_point}
Interior point methods (IPMs) solve inequality-constrained problems of the form:
\begin{subequations}
\begin{align}
\label{eq:inequality_constrained_problem}
    \min_{z} \quad &\frac{1}{2} z^\top Qz+g^\top z \\
    \text{subject to} \quad &Az + b \geq 0
\end{align}
\end{subequations}
by solving the following subproblem with slack variables $s$ and a \emph{log barrier} with relaxation parameter $\kappa > 0$ that produces an infinite penalty as $s$ approaches zero:
\begin{subequations}
\begin{align}
    \label{eq:log_barrier_ineq}
    \min_{z} \quad &\frac{1}{2} z^\top Qz+g^\top z - \kappa\bm{1}^\top  \log(s) \\
    \text{subject to} \quad &Az + b = s
\end{align}
\end{subequations}
The stationarity condition with respect to $s$ is
\begin{align}
    \lambda - \frac{\kappa}{s} = 0
\end{align}
where $\lambda$ is the Lagrange multiplier corresponding to the constraint $Ax + b = s$. Multiplying both sides by $s$ gives $s \odot \lambda = \kappa$, referred to as the relaxed complementarity condition. This paper builds on existing work in \cite{permenter_log-domain_2023}, which recognized that solutions to $s \odot \lambda = \kappa$ for positive $s$ and $\lambda$ form a smooth curve that can be parameterized by $s = \sqrt{\kappa}e^{\sigma}, \,\,\lambda=\sqrt{\kappa}e^{-\sigma}$.

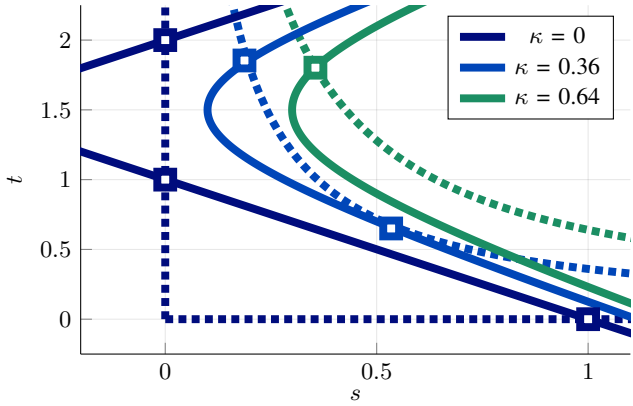
\begin{figure}[t]
    \centering

\definecolor{relaxDarkBlue}{RGB}{0.0, 12.189765, 124.51650000000001}
\definecolor{relaxBlue}{RGB}{0.0, 69.9088875, 184.30380000000002}
\definecolor{relaxDarkGreen}{RGB}{26.526375, 142.418775, 98.729625}
\definecolor{relaxGreen}{RGB}{53.914012500000005, 208.721325, 25.482149999999997}

\begin{tikzpicture}[/tikz/background rectangle/.style={fill={rgb,1:red,1.0;green,1.0;blue,1.0}, fill opacity={1.0}, draw opacity={1.0}}, show background rectangle,
comp_line/.style={line width=3pt},
comp_mark/.style={mark size={3pt}},
trim left=-30pt]
\begin{axis}[
    width=0.5\textwidth, height=0.35\textwidth,
    xlabel={$s$},
    ylabel={$t$},
    xlabel style={yshift=5pt},    
    ylabel style={yshift=-10pt},   
    xmin=-0.2, xmax=1.1,
    ymin=-0.25,  ymax=2.25,
    xtick={-0.5, 0.0, ..., 1.5},
    ytick={-0.5, 0.0, ..., 2.5},
    xmajorgrids=true,
    ymajorgrids=true,
    grid style={black, opacity=0.1},
    axis lines*=left,
    legend pos=north east,
    font=\small
]
    \addplot [color=relaxDarkBlue, comp_line, dotted, forget plot] coordinates {(0,0) (0,3)};
    \addplot[domain=0.0:1.5, samples=50, smooth, comp_line, color=relaxDarkBlue, dotted, forget plot] {0};
    \addplot[domain=0.0:1.5, samples=50, smooth, comp_line, color=relaxBlue, dotted, forget plot] {0.36/x};
    \addplot[domain=0.0:1.5, samples=50, smooth, comp_line, color=relaxDarkGreen, dotted, forget plot] {0.64/x};

    \addplot [domain=-0.5:2.5, samples=50, smooth, comp_line, color=relaxDarkBlue, solid, variable=\t] ({(-1+((2*t - 3)^2 + 4*0.001)^0.5)*0.5}, {t});
    \addlegendentry {$\kappa$ = 0}
    \addplot [domain=-0.5:2.5, samples=100, smooth, comp_line, color=relaxBlue, solid, variable=\t] ({(-1+((2*t - 3)^2 + 4*0.36)^0.5)*0.5}, {t});
    \addlegendentry {$\kappa$ = 0.36}
    \addplot [domain=-0.5:2.5, samples=100, smooth, comp_line, color=relaxDarkGreen, solid, variable=\t] ({(-1+((2*t - 3)^2 + 4*0.64)^0.5)*0.5}, {t});
    \addlegendentry {$\kappa$ = 0.64}

    \addplot[color={rgb,1:red,1.0;green,1.0;blue,1.0}, name path={15}, only marks, draw opacity={1.0}, line width={0}, solid, mark={square*}, comp_mark, mark repeat={1}, mark options={color=relaxDarkBlue, draw opacity={1.0}, fill={rgb,1:red,1.0;green,1.0;blue,1.0}, fill opacity={1.0}, comp_line, rotate={0}, solid}, forget plot]
        table[row sep={\\}]
        {
            \\
            0 2  \\
            0 1  \\
            1 0 \\
        };
    \addplot[color={rgb,1:red,1.0;green,1.0;blue,1.0}, name path={15}, only marks, draw opacity={1.0}, line width={0}, solid, mark={square*}, comp_mark, mark repeat={1}, mark options={color=relaxBlue, draw opacity={1.0}, fill={rgb,1:red,1.0;green,1.0;blue,1.0}, fill opacity={1.0}, comp_line, rotate={0}, solid}, forget plot]
        table[row sep={\\}]
        {
            \\
            0.1873506543473577  1.8537618275551073  \\
            0.5342319306953657  0.6500987145125278  \\
        };
    \addplot[color={rgb,1:red,1.0;green,1.0;blue,1.0}, name path={18}, only marks, draw opacity={1.0}, line width={0}, solid, mark={square*}, comp_mark, mark repeat={1}, mark options={color=relaxDarkGreen, draw opacity={1.0}, fill={rgb,1:red,1.0;green,1.0;blue,1.0}, fill opacity={1.0}, comp_line, rotate={0}, solid}, forget plot]
        table[row sep={\\}]
        {
            \\
            0.35514078377697794  1.8021022344814914  \\
        };
\end{axis}
\end{tikzpicture}
    \vspace{-12mm}
    \caption{Feasible sets of $st=\kappa$ and $(s - t + 2)(s + t - 1)=\kappa$ for varying $\kappa$. Three disjoint feasible points exist at $\kappa = 0$ which reduces to one as $\kappa$ grows.}
    \label{fig:relaxation_solutions}
    \vspace{-3mm}
\end{figure}

\section{Existing Works} \label{sec:existing_approaches}
A variety of approaches to the more general class of mathematical programs with complementarity constraints exist \cite{luo_mathematical_1996, raghunathan_interior_2005} and fall into three main categories: mixed-integer reformulations, smoothing or relaxation methods, and penalty methods. 

\subsubsection{Mixed-Integer Reformulation} Complementarity constraints \eqref{eq:complementarity_definition} may be re-formulated using the big-$M$ method with bounds $m_s \geq s, m_t \geq t$ and binary variable $\delta \in \{0, 1 \}$ \cite{hall_lcqpow_2025}:
\begin{align}
    \label{eq:MIP_comp_reformulation}
    0 \leq s \perp t \geq 0 \iff 
    \begin{cases}
        0 \leq s \leq m_s \delta \\
        0 \leq t \leq m_t (1 - \delta) \\
    \end{cases}
\end{align}
The resulting mixed integer programs are often solved with branch-and-bound methods implemented in commercial solvers like Gurobi \cite{gurobi_optimization_llc_gurobi_2026}. Branch-and-bound methods scale combinatorially in the number of integer variables, leading to prohibitively long solution times.

\subsubsection{Smoothing and Relaxation Methods} \label{sec:relaxation} For a general review of smoothing and relaxation methods, including nonlinear program (NLP) reformulations, we refer to \cite{fletcher_solving_2004-1}. This work uses the relaxation introduced in \ref{sec:interior_point}. Typically, methods in this category perform \emph{continuation}, where a parameter is gradually varied to approach the true non-smooth solution. One challenge encountered with relaxation methods is that they can change the topology of the feasible set associated with complementarity constraints, unlike in interior-point methods for inequalities. For example, Figure \ref{fig:relaxation_solutions} shows two overlapping complementarity constraints: $s \cdot t=\kappa$ and $(s - t + 2)\cdot (s + t - 1)=\kappa$ where, as $\kappa$ increases, the number of solutions changes from three to one. This can bias the solution towards particular local minima, but is not inherently limiting provided that the relaxed feasible set approaches the original one as $\kappa$ approaches zero.

\subsubsection{Penalty Methods} In penalty-based methods, constraints are incorporated through penalty terms in the cost function to avoid issues like LICQ. Similar to relaxation methods, continuation is used to solve subproblems for increasing penalty values. LCQPow, an open-source software package for LCQPs, includes $s \odot t = 0$ as a quadratic penalty $\rho \cdot s^\top t$, leaving the inequality constraints explicit, which forms a QP subproblem that can be solved with off-the-shelf solvers. This method suffers from two weaknesses: First, if the problem violates a constraint qualification called MPEC-LICQ described in \cite{hall_lcqpow_2025}, LCQPow often encounters failures that depend heavily on the lower-level QP solver since the subproblem solution is not unique. Second, the solver can stall at infeasible stationary points where both the quadratic cost and complementarity penalty are orthogonal to the boundary of the feasible region.

\section{Complementarity by Construction}\label{sec:coco}

\begin{figure}[t] 
    \centering
    \definecolor{relaxGrey}{RGB}{120, 120, 120}
\definecolor{relaxBlack}{RGB}{0, 0, 0}

\begin{tikzpicture}[/tikz/background rectangle/.style={fill={rgb,1:red,1.0;green,1.0;blue,1.0}, fill opacity={1.0}, draw opacity={1.0}}, show background rectangle,
comp_line/.style={line width=2pt},
trim left=-30pt]
\begin{axis}[
    width=0.50\textwidth, height=0.3\textwidth,
    xlabel={$s$},
    ylabel={$t$},
    xlabel style={yshift=5pt},    
    label style={font=\small},
    tick label style={font=\small},
    xmin=-0.1, xmax=1,
    ymin=-0.05,  ymax=1.1,
    xtick={-0.5, 0.0, ..., 1.5},
    ytick={-0.5, 0.0, ..., 2.5},
    xmajorgrids=true,
    ymajorgrids=true,
    grid style={black, opacity=0.1},
    axis lines*=left,
    legend pos=north east,
]
\addplot[color={rgb,1:red,0.8;green,0.0;blue,0.0}, name path={12}, only marks, draw opacity={1.0}, line width={0}, solid, mark={*}, mark size={3pt}, mark repeat={1}, mark options={color={rgb,1:red,0.9;green,0.0;blue,0.0}, draw opacity={1.0}, fill={rgb,1:red,0.9;green,0.0;blue,0.0}, fill opacity={1.0}, rotate={0}, solid}, forget plot]
        table[row sep={\\}]
        {
            \\
            0.346 0.346  \\
        }
        ;
    \addplot[
        domain=0.268:0.5, 
        samples=50, 
        smooth, 
        thick, 
        color={rgb,1:red,0.9;green,0.0;blue,0.0}, 
        <->,
        forget plot
    ] {0.15/x} node[pos=0.5, anchor=south west] {$\sigma$};

    \addplot [color=relaxGrey, comp_line, solid, forget plot] coordinates {(0,0) (0,3)};
    \addplot[domain=0.0:1.5, samples=50, smooth, comp_line, color=relaxGrey, solid, forget plot] {0};
    \addplot[domain=0.0:1.5, samples=100, smooth, comp_line, color=relaxGrey, solid, forget plot] {0.0225/x};
    \addplot[domain=0.0:1.5, samples=50, smooth, comp_line, color=relaxBlack, solid, forget plot] {0.1225/x};
    \addplot[domain=0.0:1.5, samples=50, smooth, comp_line, color=relaxGrey, solid, forget plot] {0.3025/x};

\end{axis}
\end{tikzpicture}
    \vspace{-12mm}
    \caption{Relaxed complementarity feasible sets for different relaxation parameters. $\sigma$ smoothly and globally parametrizes a feasible set for a fixed relaxation (black, red dot).}
    \label{fig:retraction}
    \vspace{-5mm}
\end{figure}
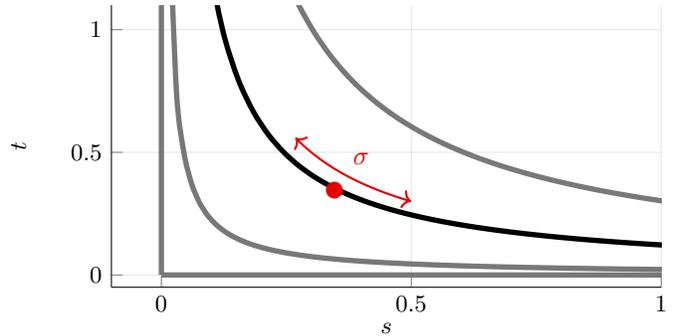

Our approach avoids the challenges associated with complementarity constraints by performing on-manifold optimization. The solution set defined by the complementarity constraints $0 \leq s \perp t \geq 0$ for scalars $s$ and $t$ is a one-dimensional continuous curve that is non-smooth at $s = t = 0$ and, therefore, does not admit a smooth parameterization. However, relaxing the constraint by replacing $s \odot t = 0$ with $s \odot t = \kappa$ for $\kappa > 0$ smooths this curve and the true constraint is recovered as $\kappa \to 0$. Importantly, the relaxed feasible set forms a Lie group: feasible points $(s, t)$ and $(p, q)$ can be composed together to create a new feasible point through scaled element-wise multiplication:
\begin{equation}
(s,t)\,\otimes\,(q,r)
:=
\left(
\frac{sq}{\sqrt{\kappa}},
\frac{tr}{\sqrt{\kappa}}
\right),
\label{eq:group_op}
\end{equation}
As first observed in \cite{permenter_log-domain_2023}, the relaxed complementarity manifold for a given $\kappa$ can now be parameterized by a single variable, $\sigma$, in the Lie algebra as shown in Figure \ref{fig:retraction}. Using the exponential map, we can map $\sigma$ back to the group using $s = \sqrt{\kappa}e^\sigma$ and $t = \sqrt{\kappa}e^{-\sigma}$. 

\subsection{Softplus Retraction Map}
While the exponential map is a natural choice, we found that it can suffer from numerical ill-conditioning and overflow errors due to its unbounded gradients.  
To address these issues, we note that any smooth function $p_\kappa(\sigma) : \mathbb{R}\to\mathbb{R^+}$ that satisfies $p_\kappa(\sigma) \cdot p_\kappa(-\sigma) = \kappa$ can be used to parameterize the feasible set, and we propose the following retraction map:
\begin{align}
    p_\kappa(\sigma) = \frac{\sqrt{\kappa}}{2}\left(\frac{\sigma}{\sqrt\kappa} + \sqrt{\left(\frac{\sigma}{\sqrt\kappa}\right)^2 + 4}\right) \label{eq:retraction}
\end{align}
This scaled \emph{softplus} function is asymptotically linear outside the corner at $\sigma = 0$, quickly approaching $p_\kappa(\sigma) = \sigma$ for positive $\sigma$ and $p_\kappa(\sigma) = 0$ for negative $\sigma$. The scaling $\frac{\sigma}{\sqrt{\kappa}}$ reflects the scaling in the group composition \eqref{eq:group_op} and makes the gradients bounded between 0 and 1. This mapping also has the convenient property that $p_\kappa(\sigma) - p_\kappa(-\sigma) = \sigma$, enabling efficient gradient and Jacobian calculations.

\section{Marble} 

We now introduce our solver algorithm, Marble.

\subsection{Problem Formulation}

Marble replaces the complementarity conditions \eqref{eq:original_complementarity_left}, \eqref{eq:original_complementarity_right}, and \eqref{eq:original_complementarity} with the implicit softplus parameterization introduced in the previous section:
\begin{subequations} \label{eq:relaxed_problem}
\begin{align}
    \min_{z, w, \sigma} \quad &\frac{1}{2} z^\top Qz + g^\top z - \kappa\bm{1}^\top \log(w) \\
    \text{subject to} \quad &Az + b = w  \\
    &Lz + l = p_\kappa(\sigma) \\
    &Rz + r = p_\kappa(-\sigma) ,
\end{align}
\end{subequations}
where $\kappa$ is the relaxation parameter, $p_\kappa(\sigma)$ is our retraction function \eqref{eq:retraction}, and inequalities are enforced through a log-barrier. For brevity in the rest of our derivation, we represent the implicit complementarity constraints as $Jz + c = h(\sigma)$, where:
\begin{align}
    J = \begin{bmatrix}
        L \\ R
    \end{bmatrix},
    \quad
    c = \begin{bmatrix}
        l \\ r
    \end{bmatrix},
    \quad
    h(\sigma) = \begin{bmatrix}
        p_\kappa(\sigma) \\
        p_\kappa(-\sigma)
    \end{bmatrix}
\end{align}
We formulate and solve this problem using an augmented Lagrangian (AL). AL methods introduce a penalty term and use multiplier estimates to ensure convergence for a finite penalty parameter \cite{nocedal_numerical_2006}. This gracefully handles infeasible subproblems and doesn't require LICQ since the penalty regularizes the multipliers \cite{izmailov_global_2012}. The corresponding augmented Lagrangian for \eqref{eq:relaxed_problem} is:
\begin{equation} \label{eq:AL_lagrangian}
\begin{aligned}
&\mathcal{L}_A(z, w, \sigma; \kappa, \rho, \alpha, \beta)
= \frac12 z^\top  Q z + g^\top  z - \kappa \bm{1}^\top  \log(w) \\
&+
\begin{bmatrix}
\alpha \\ \beta
\end{bmatrix}^\top 
\begin{bmatrix}
Az + b - w \\
Jz + c - h(\sigma)
\end{bmatrix}
+
\frac\rho2
\left\lVert
\begin{bmatrix}
Az + b - w \\
Jz + c - h(\sigma)
\end{bmatrix}
\right\rVert_2^2 ,
\end{aligned}
\end{equation}
where $\alpha$ and $\beta$ are Lagrange multiplier estimates. Our solver consists of an inner loop that minimizes $\mathcal{L}_A$ for fixed $\kappa, \rho, \alpha, \beta$ and an outer loop that updates $\kappa, \rho, \alpha, \beta$ as described in Section \ref{sec:solver_strategy}.

We use Newton's method to minimize \eqref{eq:AL_lagrangian} in the inner loop. We derive the KKT system below, introducing substitutions to improve its conditioning. The KKT conditions are:
\begin{subequations}
\begin{align}
    \frac{\partial \mathcal{L}_A}{\partial z} &=
    Qz + g + \begin{bmatrix} A \\ J \end{bmatrix}^\top 
    \begin{bmatrix}
        \alpha + \rho(Az + b - w)) \\
        \beta + \rho(Jz + c - h(\sigma)))
    \end{bmatrix}
    = 0
    \\
    \frac{\partial \mathcal{L}_A}{\partial w} &= 
    -\left(
        \alpha + \rho(Az + b - w)
    \right) - \frac{\kappa}{w}
    = 0
    \label{eq:stationarity_w}
    \\
    \frac{\partial \mathcal{L}_A}{\partial \sigma} &= -\left(\frac{\partial h}{\partial \sigma}\right)^\top  (\beta + \rho(Jz + c - h(\sigma))) = 0
\end{align}
\end{subequations}
The Jacobian $\partial h/\partial\sigma$ can be computed efficiently using properties of our retraction map \eqref{eq:retraction}:
\begin{align}
    \frac{\partial h}{\partial \sigma} &=
    \begin{bmatrix}
        p'_\kappa(\sigma) \\
        -p'_\kappa(-\sigma)
    \end{bmatrix}
    =
    \begin{bmatrix}
        p'_\kappa(\sigma) \\
        p'_\kappa(\sigma) - I
    \end{bmatrix}
\end{align}

This system of equations in the primal variables $z, w$ and $\sigma$ becomes ill-conditioned as the penalty $\rho$ increases. We avoid this using a standard primal-dual reformulation \cite{nocedal_numerical_2006} by introducing dual variables $\lambda_w \in \mathbb{R}^{m}$ and $\lambda_\sigma \in \mathbb{R}^{2p}$ with the following relationships:
\begin{align}
 \lambda_w  &= \alpha + \rho(Az + b - w )\\ 
 \lambda_\sigma &= \beta + \rho(Jz + c - h(\sigma))
\end{align}
Substituting $\lambda_w$ into the stationarity condition for the inequality slack $w$ \eqref{eq:stationarity_w} and multiplying both sides by the diagonal matrix $\text{diag}(w)$ results in the interior-point relaxed complementarity condition introduced in Section \ref{sec:interior_point}, $\mathrm{diag}(w) \lambda_w = -\kappa \bm{1}$. We satisfy this as in \cite{permenter_log-domain_2023} using our retraction map $p_\kappa$ with variable $v$.
\begin{align}
    w = p_\kappa(v), \,\, \lambda_w = -p_\kappa(-v)
\end{align}
We now arrive at the following set of conditions, which we refer to as the KKT residual $r$, that we drive to zero in each inner loop:
\begin{equation}
\underbrace{
\begin{bmatrix} Qz + g + A^\top \lambda_w + J^\top \lambda_\sigma
    \\
    \lambda_w + p_\kappa(-v)\\
    -\left(\frac{\partial h}{\partial \sigma}\right)^\top  \lambda_\sigma
    \\
   \rho^{-1}(\alpha - \lambda_w) + Az + b - p_\kappa(v) \\
   \rho^{-1}(\beta - \lambda_\sigma) + Jz + c - h(\sigma)\\ 
\end{bmatrix}
}_{\text{\normalsize $r(z,\,v,\,\sigma,\,\lambda_w,\,\lambda_\sigma)$}}
=
0
\label{eq:kkt_residual}
\end{equation}
This system has the following Jacobian, which we symmetrize by multiplying the second row by $-p'_\kappa(v)$ which lets us take advantage of fast linear solvers for sparse symmetric quasi-definite matrices. 
\begin{align} \label{eq:kkt_system}
\underbrace{
\begin{bmatrix}
    Q & & & A^\top  & J^\top  \\
      & -p'_\kappa(-v) & & I & \\
      &  & -\frac{\partial^2}{\partial \sigma^2}\left(h(\sigma)^\top  \lambda_\sigma\right) & & -\frac{\partial h}{\partial \sigma}^\top  \\
      A & -p'_\kappa(v) & & -\rho^{-1}I & \\
      J  & & -\frac{\partial h}{\partial \sigma} & & -\rho^{-1}I
\end{bmatrix}
}_{\text{\normalsize $\nabla r(z,\,v,\,\sigma,\,\lambda_w,\,\lambda_\sigma)$}}
\end{align}

\subsection{Solver Strategy} \label{sec:solver_strategy}

\begin{algorithm}
\caption{Marble Algorithm}\label{alg:lcqp}
\begin{algorithmic}[1]
\Require Problem data, optional initial guess for $z$
\State Initialize $\rho \leftarrow \rho_0$, $\kappa \leftarrow \kappa_0, \mathcal{F} \leftarrow \varnothing$
\State Initialize $s \leftarrow [z,\sigma,v,\lambda_w,\lambda_\sigma]$, $[\alpha , \beta] \leftarrow [0, 0]$
\For{$i = 1, 2, \ldots, N_{\max}$}
    \State Compute residual $r$, Jacobian $\nabla r$ from \eqref{eq:kkt_residual}
    \State Initialize regularizer $\delta \leftarrow 0$
    \Loop
        \State $[L, D] \leftarrow \textsc{QDLDL}(\nabla r,\, \delta)$
        \If{$D$ has incorrect inertia}
            \State $\delta \leftarrow \max(10^{2},\, 10\delta)$; \textbf{continue}
        \EndIf
        \State $\Delta s \leftarrow \textsc{SolveKKT}(L, D, r)$
        \State $s^+ \leftarrow \textsc{FilterLinesearch}(s, \Delta s)$
        \If{line-search succeeds}
            \State \textbf{break}
        \Else
            \State $\delta \leftarrow \max(10^{2},\, 10\delta)$
        \EndIf
    \EndLoop
    \State $s \leftarrow s^+$
    \If{$\lVert$ \eqref{eq:kkt_residual} $\rVert_\infty \leq \epsilon_{\mathrm{res}}$}
        \If{constraint violations $\leq \epsilon_e,\, \epsilon_i,\, \epsilon_c$}
            \State \textbf{return} $s$ \Comment{Solution found}
        \ElsIf{$\rho < \rho_{\max}$}
            \State $\rho \leftarrow \min(\gamma_{\rho}\, \rho,\, \rho_{\max})$
        \Else
            \State $\kappa \leftarrow \max(\gamma_{\kappa}\, \kappa,\, \kappa_{\min})$
            \State $[\alpha, \beta] \leftarrow [\lambda_w, \lambda_\sigma]$
        \EndIf
    \EndIf
\EndFor
\end{algorithmic}
\end{algorithm}

Given the non-convexity of our problem, we adopt standard strategies from the nonlinear programming literature. The inner problem is solved using Newton's method with a filter line search for step evaluation and inertia correction to ensure descent as in IPOPT \cite{nocedal_numerical_2006, wachter_implementation_2006}. The filter terms are similar to the ones defined in \cite{howell_calipso_2023}.  The KKT system is factorized and solved using QDLDL, a fast serial LDL factorization routine \cite{stellato_osqp_2020}. Similar to OSQP \cite{stellato_osqp_2020}, we apply a fixed scaling computed using Ruiz equilibration for each linear solve to reduce floating point error \cite{ruiz_scaling_2001}. 

The inner loop runs until the $\infty$-norm of the KKT residual defined by \eqref{eq:kkt_residual} is below a tolerance $\epsilon_{r}$ (default $10^{-6}$). In the outer loop, we first update the penalty parameter $\rho$. Once $\rho = \rho_{\max}$, we update $\kappa$ geometrically with a scalar $\gamma_{\kappa}$ (default $0.5$) and update the Lagrange multiplier estimates. We check the norm of the KKT residual and the violation of the original constraints to determine solver convergence. The full Marble algorithm is summarized in Algorithm \ref{alg:lcqp} where $s$ is the solution vector $[z,\sigma,v,\lambda_w,\lambda_\sigma]$.

\begin{table}[t] 
    \centering
    \caption{Solver parameters and their default values.}
    \label{tab:solver_parameters}
    \begin{tabular}{l l l}
        \toprule
        \textbf{Name} & \textbf{Symbol} & \textbf{Value} \\
        \midrule
        Penalty Initial      & $\rho_0$          & $10$            \\
        Penalty Scaling      & $\gamma_{\rho}$    & $10$ \\
        Penalty Max         & $\rho_{\mathrm{max}}$      & $10^7$\\
        Relaxation Initial   & $\kappa_0$         & $0.1$ \\
        Relaxation Scaling   & $\gamma_{\kappa}$  & $0.5$ \\
        Relaxation Min       & $\kappa_{\mathrm{min}}$    & $10^{-9}$ \\
        Residual Tol.   & $\epsilon_{\mathrm{res}}$ & $10^{-6}$ \\
        Equality Tol. & $\epsilon_e$     & $10^{-8}$ \\
        Inequality Tol. & $\epsilon_i$     & $10^{-8}$ \\
        Complementarity Tol. & $\epsilon_c$     & $10^{-8}$ \\
        \bottomrule
    \end{tabular}
\end{table}
\begin{figure}[t]
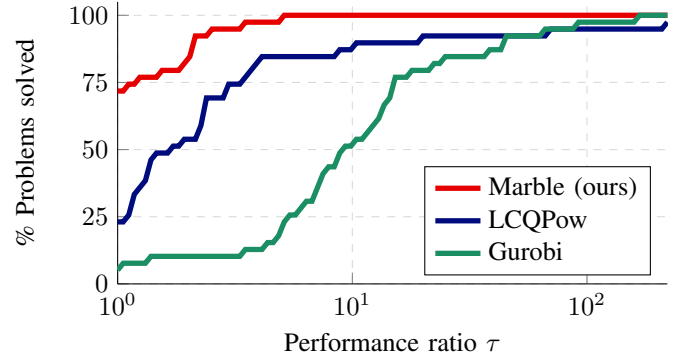
 
    \centering
    \include{figs/performance_profile}
    \vspace{-10mm}
    \caption{Total problems solved versus performance ratio (solve time scaled by minimum time per problem) for each solver. Marble is the fastest for 72\% of problems and is at most 4.85x slower.}
    \label{fig:performance_profile}
    \vspace{-5mm}
\end{figure}

\section{Experiments} \label{sec:experiments}
The Marble implementation and code to reproduce the following results are made publicly available 
\ifanonymous
\footnote{Code: omitted for review}.
\else
\footnote{Code: \texttt{https://roboticexplorationlab.org/Marble}}
\fi

We evaluate Marble on the standard MacMPEC benchmark suite \cite{noauthor_macmpec_nodate} and three robotics-specific problems, and compare the solutions to LCQPow, using qpOASES \cite{ferreau_qpoases_2014} as the QP solver, and Gurobi, where Gurobi provides the ground truth global solution. The same set of solver parameters was used for all problems and is listed in Table \ref{tab:solver_parameters}. Each problem was initialized with an all-zero initial guess. 

\subsection{MacMPEC Benchmark}

The MacMPEC benchmarks contains a variety of complementarity problems from fields such as game theory, operations research, and structural dynamics \cite{luo_mathematical_1996}. We solved the 39 MacMPEC problems that are LCQPs and compare our solver's performance against LCQPow \cite{hall_lcqpow_2025}, an SQP-based method using qpOASES as the underlying QP solver, and Gurobi \cite{gurobi_optimization_llc_gurobi_2026}, a mixed integer branch-and-bound-based method. 

Our solver obtains feasible solutions for all problems and finds the global solution for 38 of 39 problems. LCQPow fails to achieve a complementarity tolerance less than $10^{-5}$ for one problem and finds the global solution for 33 problems. For every problem, our method achieves an equal or better solution compared to LCQPow. We compare solution times using performance ratios $\tau$ for each problem, which is the solve time per solver divided by the minimum time across all three solvers. Figure \ref{fig:performance_profile} plots the percent of problems solved with at most performance ratio $\tau$ for each solver, showing that Marble is the fastest for 69\% of problems and is at most eight time slower.

\begin{table*}[!t]
\centering
\caption{Solver comparison across robotics tasks.}
\label{tab:solver_comparison_3x3}
\small
\setlength{\tabcolsep}{6pt}
\renewcommand{\arraystretch}{1.15}

\begin{threeparttable}


\begin{adjustbox}{width=\textwidth}
\begin{tabular}{l ccc @{\hspace{12pt}} ccc @{\hspace{12pt}} ccc}
\toprule
& \multicolumn{3}{c}{\makecell{\textbf{Contact Dynamics}\\[-1pt]
\footnotesize $n_z = 1560,\ n_e= 780, n_i=540,\ n_c=540$}}
& \multicolumn{3}{c}{\makecell{\textbf{Progress Constraints}\\[-1pt]
\footnotesize $n_z = 1706,\ n_e= 760, n_i=1942,\ n_c=120$}}
& \multicolumn{3}{c}{\makecell{\textbf{State-triggered Constraints}\\[-1pt]
\footnotesize $n_z = 678,\ n_e= 444, n_i=718,\ n_c=40$}} \\
\cmidrule(lr){2-4}\cmidrule(lr){5-7}\cmidrule(lr){8-10}
\textbf{Solver}

& \makecell{Feasibility} & \makecell{Objective} & \makecell{Solve Time}
& \makecell{Feasibility} & \makecell{Objective} & \makecell{Solve Time}
& \makecell{Feasibility} & \makecell{Objective} & \makecell{Solve Time} \\
\midrule

Gurobi \cite{gurobi_optimization_llc_gurobi_2026}
& \cmark & $29.20$ & $344.56$ s
& \cmark & $464.46^*$ & 845 ms
& \cmark & $380.71^*$ & $45$ ms \\

LCQPow \cite{hall_lcqpow_2025}
& \cmark & $29.21$ & 55.11 s
& \xmark & \na & \na
& \xmark & \na & \na \\

Marble (ours)
& \cmark & $31.68$ & 130 ms
& \cmark & $465.56$ & 69 ms
& \cmark & $381.19$ & 34 ms \\

\bottomrule
\end{tabular}
\end{adjustbox}

\begin{tablenotes}[flushleft]\footnotesize
\item $^*$ Indicates Gurobi achieves global optimum. 
\end{tablenotes}
\vspace{-7mm}
\end{threeparttable}
\end{table*}

\subsection{Robotics Benchmarks} \label{sec:robo_benchmarks}
We formulate and solve three robotics-specific problems chosen to demonstrate the capabilities of LCQPs to model a wide variety of systems and behaviors. The feasibility, objective and solve times for each solver on each problem is shown in Table \ref{tab:solver_comparison_3x3}. Marble is faster than LCQPow and Gurobi for all problems and is able to solve both problems LCQPow fails on.

\subsubsection{Contact} 
We solve a trajectory optimization problem for a planar hopper traversing a raised platform with stairs as shown in Figure \ref{fig:top_figure}. Making and breaking contact is modeled as complementarity between the signed distance to the floor $d$ and the normal force $f_n$, avoiding force-at-a-distance or penetration artifacts. The hopper exerts a friction force against the ground, and complementarity constraints on the tangential velocity prevent slipping when in contact. Using $v_t$ as the tangential velocity, this is expressed as:
\begin{align}
    d &\geq 0 \\
    v_t &= v_t^+ - v_t^- \\
    v_t^+,\; v_t^- &\geq 0 \\
    0 \leq v_t^+ + v_t^- + d \;&\perp\; f_n \geq 0 \label{eq:hopper_comp_v_d} \\
    -\mu f_n \leq f_f &\leq \mu f_n
\end{align}
where the positive slack variables $v_t^+, v_t^-$ upper-bound the absolute value of the tangential velocity through $v_t^+ + v_t^-$ which must be zero when the normal force is non-zero to prevent slip. 
The last constraint on the friction force $f_f$ represents the planar friction cone.

We compute the signed distance using $d = z_{\mathrm{foot}} - h(x_{\mathrm{foot}})$ where $h(x_{\mathrm{foot}})$ is the height map of the stairs, represented as the sum of four shifted and scaled sign functions. Sign functions cannot be explicitly used in an LCQP, but can be implicitly represented through a linear program:
\begin{align}
s = \operatorname{sgn}(x) = \argmin_{-1 \le s \le 1} (-sx).
\end{align}
We use the KKT conditions of the linear program as constraints in the LCQP with additional decision variables for $s$ and the Lagrange multipliers $\lambda^+$, and $ \lambda^-$. Since the KKT conditions are necessary and sufficient, $s = \operatorname{sgn}(x)$ at a feasible solution to the LCQP.

The hopper dynamics are linearized and represented as two point masses for the head and foot, where the state is their cartesian positions and velocities and the controls are prismatic forces between them. We use a quadratic tracking cost on the $x$ position of both bodies, corresponding to translating with a fixed speed, and a small quadratic cost on the controls and their first derivatives. An additional quadratic cost penalizes the vertical distance between the two point masses.

\begin{figure}[t]
    \centering
    \includegraphics[width=\linewidth]{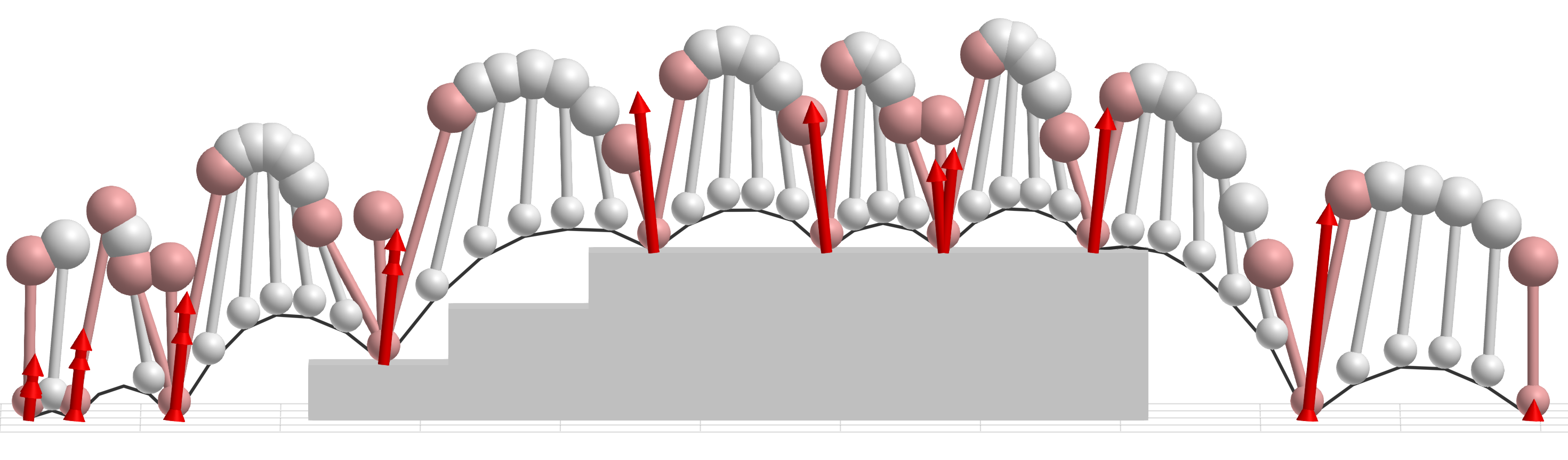}
    \caption{Hopper trajectory with contact forces (red arrows). The hopper is shaded red to indicate contact (signed distance $\leq 10^{-4}$ m.}
    \label{fig:hopper_forces}
    \vspace{-5mm}
\end{figure}

Figure \ref{fig:hopper_forces} shows the solved trajectory for the hopper problem with a regular hopping gait and hard contact events with forces that satisfy the friction cone. LCQPow, Gurobi, and Marble all achieve a feasible, locally optimal solution for this problem. Notably, Marble achieves a solution within 10\% of LCQPow and Gurobi in orders of magnitude less time. Gurobi struggles to find a global optimum since many feasible points achieve near-identical objective values. 


\subsubsection{State-Triggered Constraints} 
State-triggered constraints (STCs) specify constraints $h(z) \geq 0$ that are only applied when a \emph{trigger condition} $g(z) > 0$ is active. This can be expressed as the implication $g(z) > 0 \implies h(z) \geq 0$ and is transcribed as:
\begin{align}
    &g^+ - g^- = g(z), \,\,\,\,\, g^+, g^- \geq 0 \label{eq:stc_eq1} \\
    &h^+ - h^- = h_i(z), \,\, h^+, h^- \geq 0 \label{eq:stc_eq2} \\
    &0 \leq g^+ \perp h^- \geq 0 \label{eq:stc_comp}
\end{align}
The trigger condition can activate multiple constraints $h_1, \ldots, h_n$ at once by modifying \eqref{eq:stc_comp} to be:
\begin{align}
    0 \leq g^+ \perp \sum_{i=1}^n h_i^- \geq 0 \label{eq:stc_conjunction}
\end{align}
We demonstrate STCs in a trajectory optimization problem to guide a rocket into a catch tower located at $x=0$ with catch arms at a height of $h_\mathrm{catch}$ with length $\ell_\mathrm{arms}$. The rocket dynamics are planar and linearized about the upright state $\theta = 0$ with unit thrust-to-weight ratio, and the controls are thrust magnitude and engine gimbal deflection angle $\alpha$. The STCs specify that when the rocket is within a specified range of the tower, both ends of the rocket must be in front of the tower with the engine pointed away from the tower to avoid plume impingement; these are written as:
\begin{align}
    g(y) &= h_\mathrm{trigger} - y \\
    h_1(p, \theta) &= \bm{e}_1^\top \left(
    p + R(\theta)
    \begin{bmatrix}
        0 \\ \ell
    \end{bmatrix}
    \right) \\
    h_2(p, \theta) &= \bm{e}_1^\top \left(
    p + R(\theta)
    \begin{bmatrix}
        0 \\ -\ell
    \end{bmatrix}
    \right) \\
    h_3(\theta, \alpha) &= \theta + \alpha ,
\end{align}
where $h_\mathrm{trigger}$ is the altitude at which the constraints become active, $\ell$ is the half-length of the rocket, $p = [x, y]$ is the rocket position, and $R(\theta)$ is a linearized body-to-world rotation matrix. These trigger and constraint functions are enforced using \eqref{eq:stc_eq1}, \eqref{eq:stc_eq2}, \eqref{eq:stc_conjunction}.

The trigger region and catch trajectory are shown in \ref{fig:top_figure}. LCQPow fails to converge on this problem and encounters infeasibility within the underlying QP solver, while Marble finds a near-optimal solution compared to the global Gurobi solution.

\subsubsection{Progress Constraints} We plan a trajectory for a quadrotor linearized about hover flying through a series of $m$ rectangular gates in order, starting from an initial configuration. Gate completion is modeled with progress state $\gamma_k \in \mathbb{R}^m$ indicating whether a gate has been passed through, and progress control variables $\mu_k \in \mathbb{R}^m$ that update $\gamma_k$ according to:
\begin{align}
    \gamma_k &= \begin{cases}
       \mu_k & k = 1  \\
       \gamma_{k-1} + \mu_k & 1< k \leq N
    \end{cases}
\end{align}
To ensure that all gates are completed in order by the end of the trajectory, we enforce:
\begin{align}
    \gamma^{(j)}_N &= 1 \quad  &&1 \leq j \leq m \\
    \gamma_k^{(j)} &\geq \gamma_k^{(j+1)} \quad
    &&1 \leq j < m, \quad1 \leq k \leq N
\end{align}
The progress control $\mu_k^{(j)}$ for gate $j$ at timestep $k$ may only be active when the drone passes through the bounds of the gate:
\begin{align}
    0 \leq \mu_k^{(j)} \perp \left\lVert R_W^j \left( r_k - o^{(j)} \right) + s_k^{(j)} \right\rVert_1 \geq 0 \label{eq:gate_comp} \\
    -\ell^{(j)} \leq s_k^{(j)} \le \ell^{(j)} ,
\end{align}
where $\ell^{(j)}$, $o^{(j)}$, and $R_W^j$ are the half-extents, position, and orientation of gate $j$, respectively. We refer the interested reader to \cite{foehn_time-optimal_2021} for interpretation of these constraints. The $L^1$ norm in \eqref{eq:gate_comp} is reformulated in terms of slack variables and a set of linear equality and inequality constraints in order to fit within the LCQP framework.

\begin{figure}[t]
    \centering
    \includegraphics[width=\linewidth]{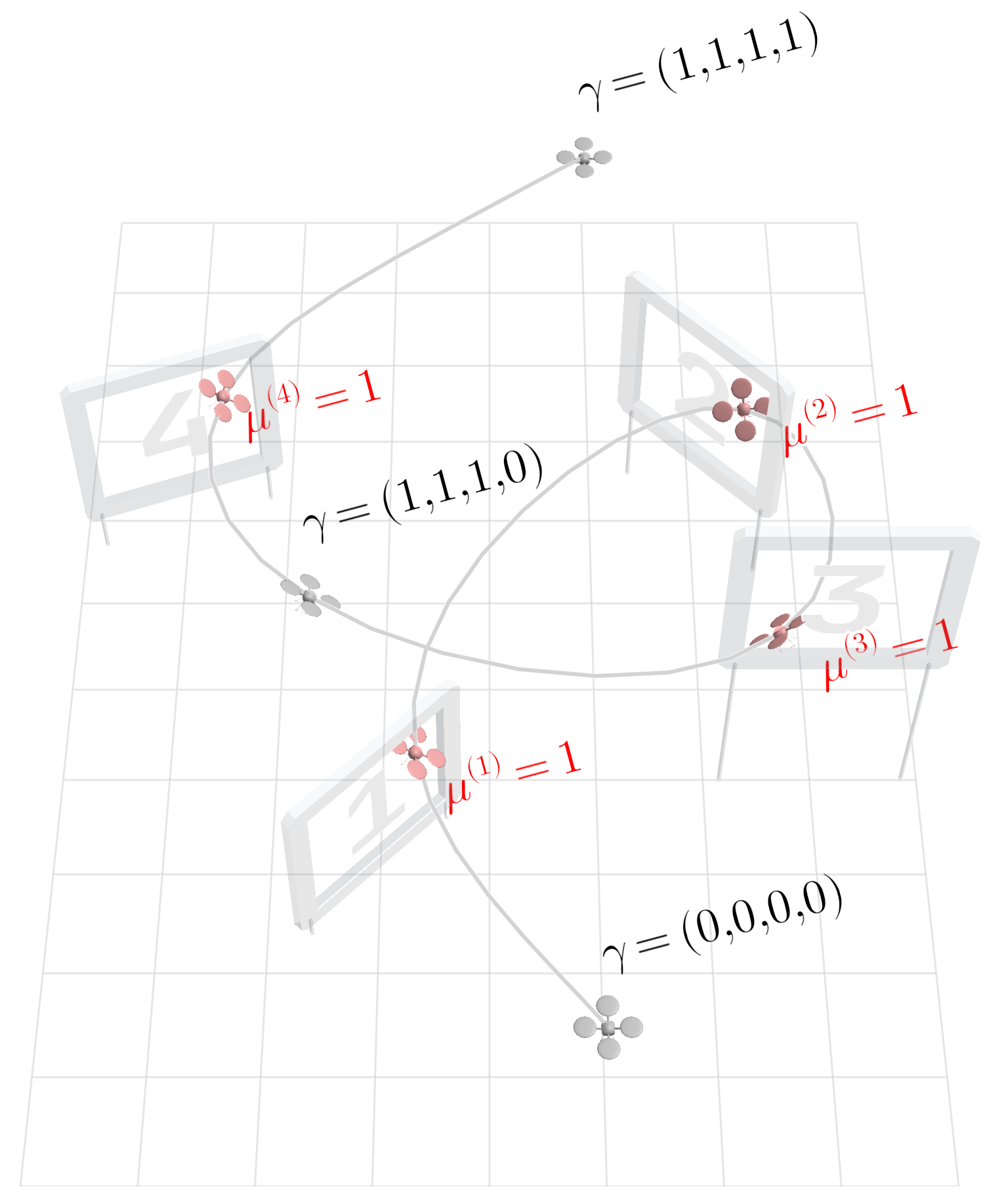}
    \caption{Gate traversal trajectory with progress constraints. The quadrotor is shaded red to indicate timesteps where the progress control variable $\mu^{(j)}$ is equal to one for gate $1 \leq j \leq 4$.}
    \label{fig:drone_gate_trigger}
    \vspace{-5mm}
\end{figure}

Figure \ref{fig:drone_gate_trigger} shows the solution to this problem which exhibits binary progress controls: $\mu_k^{(j)} = 1$ is satisfied for exactly one timestep per gate when the quadrotor flies through gate $j$ and is zero otherwise throughout the trajectory. LCQPow again fails to converge on this problem and encounters infeasibility within the underlying QP solver, while Marble finds a near-optimal solution compared to the global Gurobi solution.

\section{Conclusions, Limitations, and Future Work} \label{sec:conclusions}
In this work, we develop an approach to solving LCQPs that leverages the Lie group structure of relaxed complementarity constraints. We introduce a softplus retraction map that offers numerical advantages over the common exponential map, and demonstrate that the resulting solver is competitive across both standard benchmarks and robotics-specific problems. Our open-source solver, Marble, is implemented in C++ with Python and Julia bindings.

LCQPs remain a challenging problem class that requires careful modeling. One limitation of our solver is the phenomenon discussed in Fig. \ref{fig:relaxation_solutions}, where relaxation can change the feasible set of a problem. This can occur, for example, in the hopper problem shown in Section \ref{sec:robo_benchmarks}: modeling the tangential velocity and distance complementarity separately, as opposed to the approach used in \eqref{eq:hopper_comp_v_d}, can lead to solver failures. More work needs to be done to characterize the failure modes and robustness properties of this approach.

There are several directions for future work: first, Marble is differentiable thanks to its relaxation strategy. These derivatives could be used in downstream learning or planning applications. Second, effective warm-starting strategies may be explored via learning. Finally, globalization strategies for challenging large-scale problems could be developed through sampling or learning approaches.

\bibliographystyle{IEEEtran}
\bibliography{references}

\end{document}